# Deep Learning on Radar Centric 3D Object Detection

Seungjun Lee[#1]

[#]Seoul National University, Korea
[*]AI COLLEGE, Korea
[1]lsjj096@snu.ac.kr

*Abstract* — Even though many existing 3D object detection algorithms rely mostly on camera and LiDAR, camera and LiDAR are prone to be affected by harsh weather and lighting conditions. On the other hand, radar is resistant to such conditions. However, research has found only recently to apply deep neural networks on radar data. In this paper, we introduce a deep learning approach to 3D object detection with radar only. To the best of our knowledge, we are the first ones to demonstrate a deep learning-based 3D object detection model with radar only that was trained on the public radar dataset. To overcome the lack of radar labeled data, we propose a novel way of making use of abundant LiDAR data by transforming it into radar-like point cloud data and aggressive radar augmentation techniques.

## I. Introduction

Due to its broad Real-World applications such as autonomous driving and robotics, the proper use of 3D object detection is one of the most crucial and indispensable problems to solve. Object detection is the task of recognizing and localizing multiple objects in a scene. Objects are usually recognized by estimating a classification probability and localized with bounding boxes. In autonomous driving, the main concern is to perform 3D object detection with accuracy, robustness and real-time. Therefore, it makes almost all the autonomous vehicles equipped with multiple sensors of multiple modalities to ensure safety: camera, LiDAR (light detection and ranging), and Radar (radio detection and ranging).

Currently, with cameras, the most widely adapted vision sensor to carry out 3D object detection is LiDAR which outputs spatially accurate 3D point clouds of its surroundings. While recent 2D object detection algorithms are capable of handling large variations in RGB images, 3D point clouds are special in the sense that their unordered, sparse and locality sensitive characteristics still show great challenges to solve 3D object detection problems. Furthermore, cameras and LiDARs are prone to harsh weather conditions like rain, snow, fog or dust and illumination.

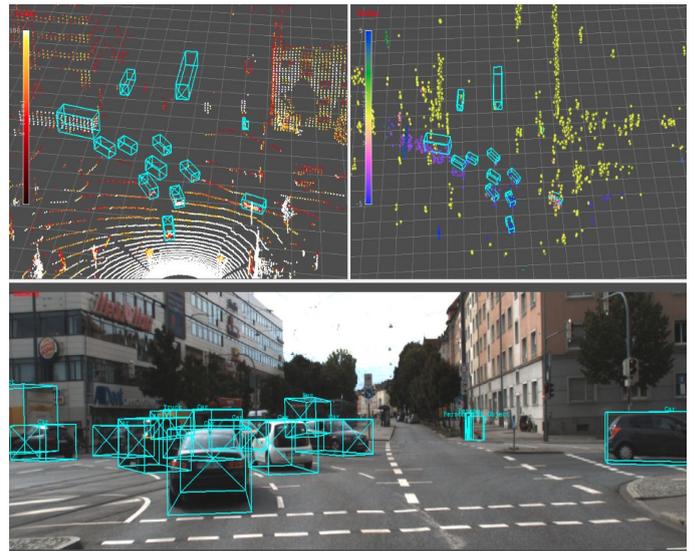

Fig. 1. An Example of radar image (up right) with the corresponding RGB camera images (down) and LiDAR images (up left) from [10].

On the contrary, automotive radar, being considerably cheaper than a LiDAR and resistant to adverse weather and insensitive to lighting variations that provides long and accurate range measurements of the surroundings simultaneously with relative radial velocity measurements by Doppler effect, is widely used within modern advanced driver assistance and vehicle safety systems. Moreover, the recent demand for autonomous radar introduced a new generation of high-resolution automotive "imaging" radar like [10] which is expected to be a substitute for expensive LiDARs.

However, there exist more difficulties in the development of radar-based detectors than LiDAR-based ones. As deep learning is a heavily data-driven approach, the top bottleneck in radar-based applications is the availability of publicly usable data annotated with ground truth information. Only the very recent nuScenes dataset [8] provides non-disclosed type of 2D radar sensor with sparsely populated 2D points but without the sampled radar ADC data required for deep radar detection whereas Astyx HiRes2019 Dataset [10] provides 3D imaging radar data that contains only 546 frames with ground-truth labels, which is relatively too small for common image datasets in the computer vision community.

Even though recent publications have shown that the radar-camera fusion object detector that exploits both images and point cloud data can be reliable to some degree in the work of [11], many things are still unknown as the research on applying deep neural networks on radar data has just begun.

To the best of our knowledge, [1] is the first and currently only one who has implemented radar-based deep neural network object detection with reliable results. But [1] was restricted to highway automotive environments and the radar-based detector has not yet been evaluated on public automotive radar datasets like [8], [10]. As some radar sensors generate 3D point clouds like LiDAR does, one may apply similar networks of LiDAR. However, the problem with this naive approach is that the point cloud has completely different properties. That is, radar data are much noisier and less accurate than LiDAR. So, it is not clear if network architectures similar to the ones used for LiDAR data are suitable for radar data.

Our contributions are the following :
1. This paper introduces a novel way of making use of abundant LiDAR point cloud data by changing their representation to radar-like point cloud data.
2. This paper introduces novel augmentation techniques for radar-only learning problems that increase the convergence speed and performance.
3. This paper demonstrates the radar-only system's viability with various state-of-the-art point cloud object detection networks.

## II. METHODS

### A. Dataset

The main dataset to evaluate is Astyx HIRES 2019's radar dataset [10] which includes a total of 546 frames and about 1,000 - 10,000 points per frame. For training, we randomly split the dataset into train, validation and test data using a ratio of 7:1.5:1.5. For evaluation, the ground truth data are split into three categories − Easy, Moderate, and Hard. For the latter, all objects are evaluated, whereas for Moderate fully occluded objects are excluded, and for Easy only fully visible objects are evaluated.

### B. Pre-training on Auxiliary task

While the largest dataset with ground-truth labels that we have reviewed is the nuScenes Dataset [8] which contains nearly 1.4M frames, with less than 500 training point clouds, training radar-based deep neural network from scratch will inevitably suffer from over-fitting. To reduce the over-fitting issue, we make use of LiDAR dataset (e.g. [7], [8], [9], [16]) to train a first neural network and reuse the first neural network's lower layers as a feature detector by the second neural network. However, when evaluated on Complex-YOLO network [2] to Astyx radar data with pre-trained weight by KITTI LiDAR dataset [7], it showed AP of 0.015% 0.034%, 1.447% for occluded, partially occluded and not occluded cars respectively which indicates that LiDAR and radar point clouds have very different properties. To enhance this method, we apply transformations on LiDAR point clouds in order to transform them into radar-like point clouds.

### C. Data Augmentations

Moreover, we employed aggressive data augmentations for point clouds and bounding box labels as below.

- Flip-x
- Global rotation-z
- Global translation-y
- Sample drop
- Global noise
- Point perturbation with Gaussian noise
- Rotate perturbation
- Jitter perturbation
- Sample Ground Truths from the Database
- Flip-y
- Global translation-x
- Random scaling
- Object Noise

The above list includes standard data augmentation during training point cloud data such as random flipping, global scaling with scaling factor uniformly sampled from [0.95, 1.05], global rotation around the vertical axis by an angle uniformly sampled from $[-\pi/4, \pi/4]$ and novel approaches were used [5], [12], [15], [17].

### D. Networks

For networks, we applied the work of state-of-the-art 3D object detection networks that takes input point clouds only, which can be divided into three methods that leverage the mature 2D detection frameworks by projecting the point clouds into bird's eye view[2], or transform the point clouds into regular 3D voxels [5], [12], and learn directly from the 3D point cloud data for point cloud classification and segmentation [13], [14].

## III. Results

For evaluation, the average precision (AP) is utilized which is a common evaluation metric for 3D object detection [7] and 3D intersection over union (IoU) threshold of 0.5 was used.

The work of [11], which achieved an average precision (AP) of (0.61, 0.48, 0.45) for the detection of cars using radar-camera fusion with Astyx radar dataset [10], was used for the baseline model. Here we divided the dataset into three difficulty categories (*Easy, Moderate, Hard*) based on the visibility/occlusion of the object.

### A. Complex-YOLO

For bird's eye view detection network, we exploit Complex-YOLO [2], a state of the art real-time one stage 3D object detection network. For the detection of cars, we crop the point cloud based on the ground-truth distribution at $[-2, 4] \times [-70, 70] \times [-70, 70]$ m along the z × y × x axes with a total of 9 prior box shapes. Widths: 1.7 m. Lengths: 4.2 m,

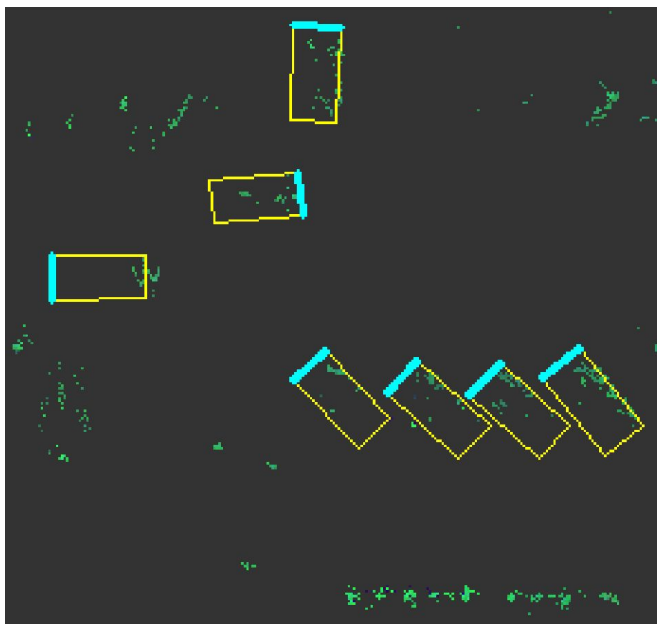

Fig. 2. An Example of bird's eye view detection on radar data.

3.85 m, 3.5 m. Orientations: 0, 1.57, -1.57 (rad).

For experiments, stochastic gradient descent was used with a momentum of 0.9 and a weight decay of 0.0005. The training consisted of 220 iterations, with batch size of 64 and an initial learning rate of 0.001.

Despite a small amount of annotated train and test data, it achieved an average precision (AP) of 0.75 for the detection of cars which is evaluated on the non-occluded objects.

## IV. Discussion

In this paper, the main question is "Can radar-only carry out object detection?" From the poor evaluation that the network pre-trained on LiDAR data showed, the representation of LiDAR and radar point cloud are very different and the naive approach of applying LiDAR based networks on radar data is likely to fail.

Considering the small dataset that we used to train, the AP of 0.75 that the network trained on radar point cloud is a decent result. Complex YOLO reported AP of (85.89, 77.40, 77.33) for the category (*Easy, Moderate, Hard*) on class *Car* which was trained on KITTI dataset that contains nearly 20 times the amount of our training data.

Moreover, according to the report on [10] that radar point clouds are about 10 times more sparse than Velodyne VLP-16 (10 Hz, 16 laser beams), the radar is able to generalize well on deep neural networks.

## V. Conclusion and Future Work

In this paper, we have presented a radar-only perception system. To the best of our knowledge, we are the first ones to demonstrate a deep-learning-based object detection model that operates on the radar-only with the public radar dataset. We proposed a novel way of handling the lack of annotated ground truth radar datasets.

Following this paper, we plan to further extend our work by evaluating more diverse state-of-the-art point cloud detection networks and introduce ways to incorporate the Doppler information to the network to enhance the radar-based network.